\documentclass[10pt,twocolumn,twoside]{IEEEtran} 

\usepackage{cite}

\usepackage{graphicx}
\ifCLASSINFOpdf
\else
\fi
\usepackage[cmex10]{amsmath}

\usepackage{mdwmath}

\usepackage[tight,footnotesize]{subfigure}

\ifCLASSOPTIONcaptionsoff
  \usepackage[nomarkers]{endfloat}
 \let\MYoriglatexcaption\caption
 \renewcommand{\caption}[2][\relax]{\MYoriglatexcaption[#2]{#2}}
\fi
\usepackage{url}

\usepackage{multirow,algorithm,algorithmic}

\begin{document}

%
\title{Robust Image Analysis by $L_1$-Norm Semi-supervised Learning}

%
%
%

\author{Zhiwu~Lu~and~Yuxin~Peng$^*$
\thanks{The authors are with the Institute of Computer Science and Technology, Peking University,
Beijing 100871, China (e-mail: luzhiwu@icst.pku.edu.cn,
pengyuxin@icst.pku.edu.cn). }
\thanks{$^*$ Corresponding author.}}

\maketitle

\begin{abstract}
This paper presents a novel $L_1$-norm semi-supervised learning
algorithm for robust image analysis by giving new $L_1$-norm
formulation of Laplacian regularization which is the key step of
graph-based semi-supervised learning. Since our $L_1$-norm Laplacian
regularization is defined directly over the eigenvectors of the
normalized Laplacian matrix, we successfully formulate
semi-supervised learning as an $L_1$-norm linear reconstruction
problem which can be effectively solved with sparse coding. By
working with only a small subset of eigenvectors, we further develop
a fast sparse coding algorithm for our $L_1$-norm semi-supervised
learning. Due to the sparsity induced by sparse coding, the proposed
algorithm can deal with the noise in the data to some extent and
thus has important applications to robust image analysis, such as
noise-robust image classification and noise reduction for visual and
textual bag-of-words (BOW) models. In particular, this paper is the
first attempt to obtain robust image representation by sparse
co-refinement of visual and textual BOW models. The experimental
results have shown the promising performance of the proposed
algorithm.
\end{abstract}



%
\IEEEpeerreviewmaketitle

\section{Introduction}

Semi-supervised learning, i.e., learning from both labeled and
unlabeled data, has been widely applied to many challenging image
analysis tasks \cite{XY09,GVS10,LI10,FWT10,TYH09,L4M09} such as
image representation, image classification, and image annotation. In
different image analysis tasks, the manual labeling of training data
is often tedious, subjective as well as expensive, while the access
to unlabeled data is much easier. Through exploiting the large
number of unlabeled data with reasonable assumptions,
semi-supervised learning \cite{BM98,ZGL03,ZBLW04,CJ05,AZ07} can
reduce the need for expensive labeled data and thus achieve
promising results especially for community-contributed image
collections (e.g. Flickr).

Among various semi-supervised learning methods, one influential work
is graph-based semi-supervised learning \cite{ZGL03,ZBLW04} which
models the entire dataset as a graph. The basic idea behind this
semi-supervised learning is label propagation on the graph with the
cluster consistency \cite{ZBLW04} (i.e. two data points on the same
geometric structure are likely to have the same class label). Since
the graph is at the heart of graph-based semi-supervised learning,
graph construction has been extensively studied
\cite{WZ08,YW09,CYY10,LHC10} in the past years. However, these graph
construction methods are not developed directly for noise reduction,
and the corresponding semi-supervised learning may suffer from
significant performance degradation due to the inaccurate labeling
of data points commonly encountered in different image analysis
tasks. For example, the annotations of images may be contributed by
the community (see Flickr) and we can only obtain noisy tags.

In this paper, we focus on proposing a novel noise-robust
graph-based semi-supervised learning method, rather than the
well-studied graph construction. As summarized in \cite{WZ08}, the
traditional graph-based semi-supervised learning can be formulated
as a quadratic optimization problem based on Laplacian
regularization \cite{ZGL03,ZBLW04,AZ07,NSZ10,FWT10}. Considering
that the sparsity induced by $L_1$-norm optimization can help to
deal with the noise in the data to some extent
\cite{Donoho04,WYG09}, if we succeed in formulating Laplacian
regularization as an $L_1$-norm term instead, we can convert the
traditional semi-supervised learning to $L_1$-norm optimization and
enable our new semi-supervised learning also to benefit from the
nice property of sparsity. Fortunately, derived from the eigenvalue
decomposition of the normalized Laplacian matrix $\mathcal{L}$, we
can readily represent $\mathcal{L}$ in a symmetrical decomposition
form, which can be further used to formulate Laplacian
regularization as an $L_1$-norm term. Since all the eigenvectors of
$\mathcal{L}$ are explored in this symmetrical decomposition, our
new $L_1$-norm Laplacian regularization can be considered to be
explicitly formulated based upon the manifold structure of the data.

As a convex optimization problem, the above $L_1$-norm
semi-supervised learning has a unique global solution. By working
only with a small subset of eigenvectors, we develop a fast sparse
coding algorithm for our $L_1$-norm semi-supervised learning. In
this paper, we only adopt the fast iterative shrinkage-thresholding
method \cite{BT09} for sparse coding, regardless of many other
$L_1$-norm optimization methods\cite{OPT00,FNW07,LBR07,DM09}. Due to
the nice property of sparsity, the proposed algorithm can deal with
the noise in the data to some extent, as shown in our later
experiments. Hence, it has important applications to robust image
analysis where noisy labels are provided. In this paper, we apply
the proposed algorithm to two typical image analysis tasks, i.e.,
noise-robust semi-supervised image classification and noise
reduction for both visual and textual bag-of-words (BOW) models.
Although only tested in these two applications, the proposed
algorithm can be extended to other image analysis tasks, given that
semi-supervised learning has been widely used in the literature.

To emphasize the main contributions of this paper, we summarize the
following distinct advantages of our novel $L_1$-norm
semi-supervised learning:
\begin{itemize}
\item
We have made the first attempt to formulate Laplacian regularization
as an $L_1$-norm term \emph{explicitly based upon the manifold
structure of the data}.
\item
Our $L_1$-norm semi-supervised learning algorithm has been shown to
\emph{achieve significant improvements in robust image analysis}
where noisy labels are provided.
\item
Our new $L_1$-norm Laplacian regularization can be \emph{similarly
applied to many other difficult problems}, considering the wide use
of Laplacian regularization.
\item
This is the first attempt to obtain robust image representation by
\emph{sparse co-refinement of visual and textual BOW models} for
community-contributed image collections.
\end{itemize}

The remainder of this paper is organized as follows.
Section~\ref{sect:rw} provides a brief review of related work. In
Section~\ref{sect:l1ssl}, we propose a fast $L_1$-norm
semi-supervised learning algorithm by defining novel $L_1$-norm
Laplacian regularization. In Section~\ref{sect:ria}, the proposed
algorithm is applied to two robust image analysis tasks:
noise-robust image classification and sparse co-refinement of visual
and textual BOW models. In Section~\ref{sect:exp}, we present the
experimental results to evaluate the proposed algorithm. Finally,
Section~\ref{sect:con} gives the conclusions.

\section{Related Work}
\label{sect:rw}

In this paper, we make attempt to formulate graph-based
semi-supervised learning as $L_1$-norm optimization so that it can
benefit from the nice property of sparsity and thus deal with the
noise in the data to some extent. This is quite different from the
attempt to construct a graph with sparse representation
\cite{YW09,TYH09,CYY10} for graph-based semi-supervised learning.
Although these two different attempts both exploit $L_1$-norm
optimization for semi-supervised learning, a new $L_1$-norm
semi-supervised learning method is proposed in the present paper
while the traditional semi-supervised learning was still used in
\cite{YW09,TYH09,CYY10}. In fact, the graph constructed with sparse
representation can be readily applied to our new $L_1$-norm
semi-supervised learning algorithm.

To formulate semi-supervised learning as an $L_1$-norm optimization
problem, we give new $L_1$-norm explanation of Laplacian
regularization explicitly based upon the manifold structure of the
data. This $L_1$-norm Laplacian regularization distinguishes our
semi-supervised learning algorithm greatly from another $L_1$-norm
semi-supervised learning algorithm proposed in \cite{SB10} which
directly adopts Lasso \cite{Tibshirani96} for semi-supervised
learning and completely ignores the important Laplacian
regularization that has been widely used for graph-based
semi-supervised learning in the literature
\cite{ZGL03,ZBLW04,AZ07,NSZ10,FWT10}. In fact, our $L_1$-norm
semi-supervised learning algorithm has been shown to outperform
\cite{SB10} significantly (see later experimental results).
Moreover, although both Laplacian regularization and $L_1$-norm
optimization have also been used in \cite{GTC10,TYH09}, the
Laplacian regularization term in the objective function is still
quadratic, which is quite different from our new $L_1$-norm
Laplacian regularization.

Since our new $L_1$-norm Laplacian regularization is defined
directly over the eigenvectors of the normalized Laplacian matrix,
we can formulate semi-supervised learning as an $L_1$-norm linear
reconstruction problem in the framework of sparse coding
\cite{Donoho04,WYG09}. Moreover, by working with only a small subset
of eigenvectors, we can develop a fast sparse coding algorithm for
our $L_1$-norm semi-supervised learning, which is efficient even for
robust image analysis tasks where the datasets are often large.
Although there exist other $L_1$-norm generalizations
\cite{CLK10,PFT11} of Laplacian regularization, they are not defined
based upon the eigenvectors and the corresponding sparse coding
algorithms incur too large time cost.

Considering the distinct advantage (i.e. noise robustness as shown
in later experiments) of our $L_1$-norm semi-supervised learning,
our original motivation is to apply it to robust image analysis
where noisy labels are provided. In particular, to our best
knowledge, we have made the first attempt to obtain robust image
representation by sparse co-refinement of visual and textual BOW
models. This strategy is extremely important for the success of
robust image analysis on community-contributed image collections
(e.g. Flickr), because it becomes rather difficult to generate
accurate visual vocabularies and obtain clean image tags in such
complicated case. However, in the literature, most previous methods
can not deal with visual and textual BOW refinement simultaneously.
For example, various supervised \cite{YJS08,MJJ10} and unsupervised
\cite{JXY09,LPI11} methods have been developed specially for visual
vocabulary optimization, while in \cite{LNR10,ZYM10,TYH09} only tag
refinement is considered for robust image analysis. More detailed
comparison to these methods can be found in
Section~\ref{sect:ria:bowr}.

\section{$L_1$-Norm Semi-Supervised Learning}
\label{sect:l1ssl}

In this section, we first give a brief review of graph-based
semi-supervised learning. To address the problem associated with
this semi-supervised learning, we further present new $L_1$-norm
formulation of Laplacian regularization. Finally, based on this
$L_1$-norm Laplacian regularization, we develop a fast $L_1$-norm
semi-supervised learning algorithm.

\subsection{Graph-Based Semi-Supervised Learning}

To introduce graph-based semi-supervised learning, we first
formulate a semi-supervised learning problem as follows. Here, we
only consider the two-class problem, while the multi-class problem
can be handled the same as \cite{ZBLW04}. Given a dataset
$\mathcal{X} = \{x_1, ..., x_l, x_{l+1},..., x_n\}$ and a label set
$\{1,-1\}$, the first $l$ data points $x_i~(i\leq l)$ are labeled as
$y_i \in \{1,-1\}$ and the remaining data points $x_u~(l+1 \leq u
\leq n)$ are unlabeled with $y_u = 0$. The goal of semi-supervised
learning is to predict the labels of the unlabeled data points,
i.e., to find a vector $\mathbf{f}=[f_1,...,f_n]^T$ corresponding to
a classification on the dataset $\mathcal{X}$ by labeling each data
point $x_i$ with a label $\mathrm{sign}(f_i)$, where
$\mathrm{sign}(\cdot)$ denotes the sign function. Let
$\mathbf{y}=[y_1,...,y_n]^T$, and we can readily observe that
$\mathbf{y}$ is exactly consistent with the initial labels according
to the decision rule.

\begin{figure*}[t]
\vspace{0.05in}
\begin{center}
\subfigure[]{
\includegraphics[height=0.24\textwidth]{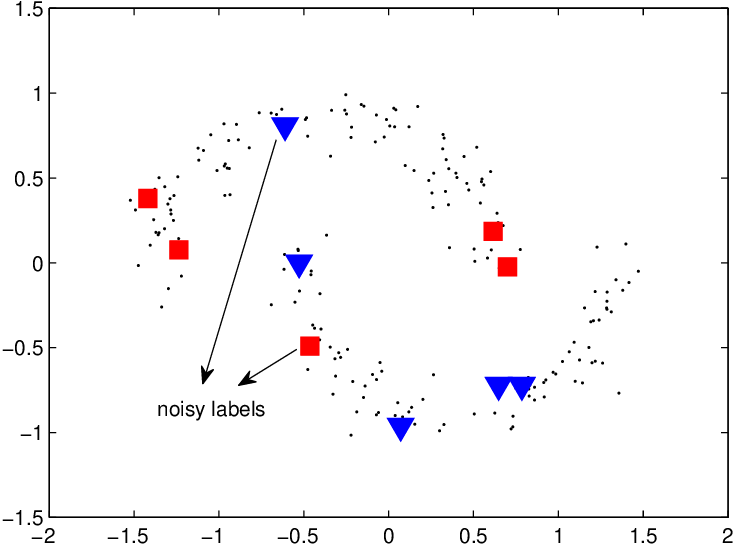}
\label{Fig.1:a}} \hspace{0.4in} \subfigure[]{
\includegraphics[height=0.24\textwidth]{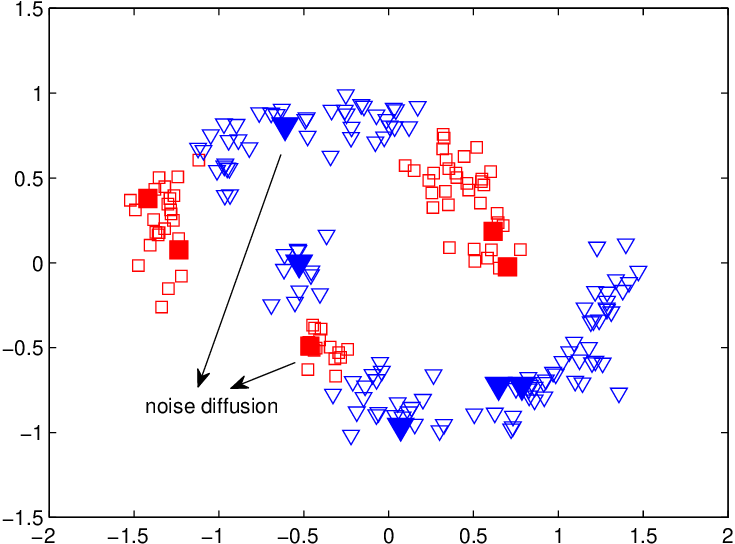}
\label{Fig.1:b}}
\end{center}
\vspace{-0.1in} \caption{(a) The two-moons toy dataset with each
class having one incorrectly (out of five) labeled data point
initially. (b) The classification results on this toy dataset by one
typical graph-based semi-supervised learning method proposed in
\cite{ZBLW04}. We can clearly observe the severe problem of noise
diffusion associated with the traditional semi-supervised learning
when noisy initial labels are provided.} \label{Fig.1}
\vspace{-0.1in}
\end{figure*}

To solve the above problem by graph-based semi-supervised learning,
we need to model the whole dataset as a graph
$\mathcal{G}=\{\mathcal{V},W\}$ with its vertex set $\mathcal{V} =
\mathcal{X}$ and weight matrix $W=[w_{ij}]_{n\times n}$, where
$w_{ij}$ denotes the similarity between $x_i$ and $x_j$. The weight
matrix $W$ is assumed to be nonnegative and symmetric. For example,
we usually define $W$ as
\begin{eqnarray}
w_{ij} = \exp(-||x_i-x_j||^2/(2\sigma^2)),
\end{eqnarray}
where the variance $\sigma$ is a free parameter that can be
determined empirically. Moreover, to eliminate the need to tune this
parameter, we can adopt the graph construction methods reported in
\cite{WZ08,YW09,CYY10}. Based on the weight matrix $W$, we compute
the normalized Laplacian matrix $\mathcal{L}$ of the graph
$\mathcal{G}$ by
\begin{eqnarray}
\mathcal{L}=I-D^{-\frac{1}{2}}WD^{-\frac{1}{2}},
\end{eqnarray}
where $I$ is an $n\times n$ identity matrix, and $D$ is an $n\times
n$ diagonal matrix with its $i$-th diagonal element being equal to
the sum of the $i$-th row of $W$ (i.e. $\sum_j w_{ij}$).

In this paper, we focus on one typical graph-based semi-supervised
learning method proposed in \cite{ZBLW04}. Its objective function
can be defined as follows:
\begin{eqnarray}
Q(\mathbf{f}) = \frac{1}{2} ||\mathbf{f}-\mathbf{y}||_2^2 +
\frac{\lambda}{2}\mathbf{f}^T\mathcal{L}\mathbf{f}, \label{eq:l2ssl}
\end{eqnarray}
where $\lambda > 0$ is a regularization parameter. Then the
classification function is given by
\begin{eqnarray}
\mathbf{f}^*=\arg \min_{\mathbf{f}} Q(\mathbf{f}).
\end{eqnarray}
The first term of $Q(\mathbf{f})$ is the fitting constraint, which
means a good classification function should not change too much from
the initial label assignment. The second term is the smoothness
constraint, which means that a good classification function should
not change too much between nearby data points. The trade-off
between these two competitive constraints is captured by the
positive parameter $\lambda$. It should be noted that the smoothness
constraint actually denotes the well-known Laplacian regularization
\cite{ZGL03,ZBLW04,AZ07,NSZ10} which has been widely used for
semi-supervised learning.

However, in the literature, the original motivation of developing
these semi-supervised learning methods is to exploit both labeled
and unlabeled data, but not to deal with the noise in the data. This
means that they are not suitable for the challenging tasks (e.g.
robust image analysis) where noisy initial labels are provided. To
clearly show this disadvantage, we give a toy example in
Fig.~\ref{Fig.1}. We can observe that the negative effect of noisy
labels is severely diffused by the traditional semi-supervised
learning. Hence, our main motivation is just to develop a new
semi-supervised learning method that can suppress the negative
effect of noisy labels. Fortunately, in the following, the problem
shown in Fig.~\ref{Fig.1} can be effectively handled by $L_1$-norm
Laplacian regularization.

\subsection{$L_1$-Norm Laplacian Regularization}
\label{sect:l1ssl:l1lap}

As reported in \cite{Donoho04,WYG09}, the sparsity induced by
$L_1$-norm optimization can help to deal with the noise in the data
to some extent. If we succeed in formulating Laplacian
regularization as an $L_1$-norm term instead, we can convert the
traditional semi-supervised learning to $L_1$-norm optimization and
enable our new semi-supervised learning also to benefit from the
nice property of sparsity (i.e. suppress the negative effect of
noisy labels). Hence, in the following, we focus on $L_1$-norm
formulation of Laplacian regularization.

Considering the important role that the normalized Laplacian matrix
$\mathcal{L}$ plays in Laplacian regularization, we first give a
symmetrical decomposition of $\mathcal{L}$. As a nonnegative
definite matrix, $\mathcal{L}$ can be decomposed into
\begin{eqnarray}
\mathcal{L}=V\Sigma V^T,
\end{eqnarray}
where $V$ is an $n\times n$ orthonormal matrix with each column
being an eigenvector of $\mathcal{L}$, and $\Sigma$ is an $n\times
n$ diagonal matrix with its diagonal element $\Sigma_{ii}$ being an
eigenvalue of $\mathcal{L}$ (sorted as $0\leq \Sigma_{11} \leq
...\leq \Sigma_{nn} $). Furthermore, we represent $\mathcal{L}$ in
the following symmetrical decomposition form:
\begin{eqnarray}
\mathcal{L}= (\Sigma^{\frac{1}{2}} V^T)^T\Sigma^{\frac{1}{2}} V^T =
B^TB,
\end{eqnarray}
where $B=\Sigma^{\frac{1}{2}}V^T$. Since $B$ is computed with all
the eigenvectors of $\mathcal{L}$, we can regard $B$ as being
explicitly defined based upon the manifold structure of the data.

We further directly utilize $B$ to define a new $L_1$-norm
smoothness measure, instead of the traditional smoothness measure
used as Laplacian regularization for semi-supervised learning. In
spectral graph theory, the smoothness of a vector $\mathbf{f} \in
R^n$ is measured by $\Omega(\mathbf{f}) = \mathbf{f}^T \mathcal{L}
\mathbf{f}$, which is exactly the smoothness constraint in equation
(\ref{eq:l2ssl}). Different from $\Omega(\mathbf{f})$, in this
paper, the $L_1$-norm smoothness of a vector $\mathbf{f} \in R^n$ is
measured by $\tilde{\Omega}(\mathbf{f}) = ||B\mathbf{f}||_1$. As for
this new $L_1$-norm smoothness measure, we have the following
proposition.
\newtheorem{prop}{Proposition}
\begin{prop}
(i) If $\tilde{\Omega}(\mathbf{f})\leq 1$, $\Omega(\mathbf{f}) \leq
\tilde{\Omega}(\mathbf{f})$; (ii) For an eigenvector $V_{.i}$ of
$\mathcal{L}$, $\tilde{\Omega}(V.i) = \Sigma^{\frac{1}{2}}_{ii}$;
(iii) If $\mathbf{f} = V \alpha = \sum_{i=1}^n \alpha_i V_{.i}$,
$\tilde{\Omega}(\mathbf{f}) = \sum_{i=1}^n |\alpha_i|
\Sigma^{\frac{1}{2}}_{ii}$.
\end{prop}

\textbf{Proof}: (i) If $\tilde{\Omega}(\mathbf{f})\leq 1$,
$\Omega(\mathbf{f}) = \mathbf{f}^T\mathcal{L}\mathbf{f} =
\mathbf{f}^TB^TB\mathbf{f} = ||B\mathbf{f}||_2^2 = \sum_{i=1}^n
(B_{i.}\mathbf{f})^2 \leq (\sum_{i=1}^n |B_{i.}\mathbf{f}|)^2 =
||B\mathbf{f}||_1^2 = (\tilde{\Omega}(\mathbf{f}))^2 \leq
\tilde{\Omega}(\mathbf{f})$, where $B_{i.}$ denotes the $i$-th row
of $B$; (ii) $\tilde{\Omega}(V.i) = ||BV_{.i}||_1 =
||\Sigma^{\frac{1}{2}}V^TV_{.i}||_1$. Since $V$ is orthonormal, we
further have $\tilde{\Omega}(V.i) = ||\Sigma^{\frac{1}{2}}
[V_{.1}^TV_{.i},..., V_{.{i-1}}^TV_{.i}, V_{.i}^TV_{.i},
V_{.{i+1}}^TV_{.i},...,V_{.n}^TV_{.i}]^T||_1 =
||[0,...,0,\Sigma^{\frac{1}{2}}_{ii},0,...,0]^T||_1 =
\Sigma^{\frac{1}{2}}_{ii}$; (iii) $\tilde{\Omega}(\mathbf{f}) =
||B\mathbf{f}||_1 = ||(\Sigma^{\frac{1}{2}}V^T)(V\alpha)||_1 =
||\sum_{i=1}^n \alpha_i \Sigma^{\frac{1}{2}}V^TV_{.i}||_1 =
||\sum_{i=1}^n \alpha_i [0,...,0, \Sigma^{\frac{1}{2}}_{ii},
0,...,0]^T||_1 =  \sum_{i=1}^n |\alpha_i|
\Sigma^{\frac{1}{2}}_{ii}$.

Proposition 1(i) shows that our $L_1$-norm smoothness can ensure the
traditional smoothness if we succeed in reducing the former below 1.
Proposition 1(ii) shows that eigenvectors with smaller eigenvalues
are smoother in terms of our $L_1$-norm smoothness measure. Since
any vector $\mathbf{f} \in R^n$ can be denoted as $\mathbf{f} = V
\alpha = \sum_{i=1}^n \alpha_i V_{.i}$, we can conclude from
Proposition 1(iii) that smooth vectors are linear combinations of
the eigenvectors with small eigenvalues.

By replacing the traditional smoothness constraint (i.e. Laplacian
regularization) in equation (\ref{eq:l2ssl}) with our $L_1$-norm
version, we define a new objective function for graph-based
semi-supervised learning as follows:
\begin{eqnarray}
\tilde{Q}(\mathbf{f}) = \frac{1}{2} ||\mathbf{f}-\mathbf{y}||_2^2 +
\lambda||B\mathbf{f}||_1. \label{eq:l1ssl}
\end{eqnarray}
The first term of $\tilde{Q}(\mathbf{f})$ is the fitting constraint,
while the second term is the $L_1$-norm smoothness constraint used
as Laplacian regularization. Here, it should be noted that the
fitting constraint is not formulated as an $L_1$-norm term. The
reason is that most elements of $\mathbf{f}$ tend to zeros (i.e.
sparsity) by $\min_{\mathbf{f}} ||\mathbf{f}-\mathbf{y}||_1 +
\lambda ||B\mathbf{f}||_1$ given that $\mathbf{y}$ has very few
nonzero elements (i.e. very few initial labeled data are often
provided for semi-supervised learning). In other words, the labels
of data points are almost not propagated across the dataset, which
completely conflicts with the original goal of semi-supervised
learning. Hence, the fitting constraint of $\tilde{Q}(\mathbf{f})$
remains as an $L_2$-norm term. In the following, $L_1$-norm
semi-supervised learning ($L_1$-SSL) refers to $\min_{\mathbf{f}}
\tilde{Q}(\mathbf{f})$.

It is worth noting that our $L_1$-norm formulation of Laplacian
regularization plays an important role in our explanation of
$L_1$-norm semi-supervised learning in the framework of sparse
coding. More concretely, according to Proposition 1(iii), our
$L_1$-norm semi-supervised learning can be formulated as a linear
reconstruction problem by setting $\mathbf{f} = V \alpha$.
Furthermore, to solve this linear reconstruction problem
efficiently, we can develop a fast sparse coding algorithm (see
Section \ref{sect:l1ssl:fsc}) by working with only a small subset of
eigenvectors (i.e. only partial columns of $V$ are used), which is
especially suitable for robust image analysis tasks where the
datasets are often large. Although there exist other $L_1$-norm
generalizations \cite{CLK10,PFT11} of Laplacian regularization which
approximately take the form of $\sum_{ij}w_{ij}|f_i-f_i|$, they are
not explicitly defined based upon the eigenvectors of $\mathcal{L}$
and the strategy of dimension reduction is hard to be used for
$\mathbf{f}$. Hence, the sparse coding algorithms developed in
\cite{CLK10,PFT11} incur too large time cost.

Finally, we can similarly utilize $B$ to formulate the traditional
Laplacian regularization as an $L_2$-norm term
\begin{eqnarray}
\mathbf{f}^T\mathcal{L}\mathbf{f}=(B\mathbf{f})^T
B\mathbf{f}=||B\mathbf{f}||_2^2,
\end{eqnarray}
which is explicitly based upon the manifold structure of the data.
Accordingly, the objective function $Q(\mathbf{f})$ of the
traditional semi-supervised learning \cite{ZBLW04} can be redefined
as the sum of two $L_2$-norm terms:
\begin{eqnarray}
Q(\mathbf{f}) = \frac{1}{2} ||\mathbf{f}-\mathbf{y}||_2^2 +
\frac{\lambda}{2}||B\mathbf{f}||_2^2.
\end{eqnarray}
In the following, the traditional semi-supervised learning
\cite{ZBLW04} is called as $L_2$-norm semi-supervised learning
($L_2$-SSL).

\subsection{Fast $L_1$-Norm Semi-Supervised Learning}
\label{sect:l1ssl:fsc}

As a convex optimization problem, our $L_1$-norm semi-supervised
learning has a unique global solution $\mathbf{f}^* =
\arg\min_{\mathbf{f}} \tilde{Q}(\mathbf{f})$. Let
$\mathbf{x}=\mathbf{f}-\mathbf{y}$, $A=B$, and
$\mathbf{b}=-B\mathbf{y}$. The original problem $\min_{\mathbf{f}}
\tilde{Q}(\mathbf{f})$ for our $L_1$-norm semi-supervised learning
is equivalently transformed into:
\begin{eqnarray}
\min_\mathbf{x} \frac{1}{2} ||\mathbf{x}||_2^2 +
\lambda||A\mathbf{x}-\mathbf{b}||_1,
\end{eqnarray}
which is a new $L_1$-norm optimization problem. Similar to
\cite{BV04}, a log-barrier algorithm can be readily developed for
this $L_1$-norm optimization. However, the obtained log-barrier
algorithm scales polynomially with the data size and then becomes
impractical for image analysis tasks.

Fortunately, as we have mentioned in Section \ref{sect:l1ssl:l1lap},
the dimension of our $L_1$-norm semi-supervised learning can be
reduced dramatically by working only with a small subset of
eigenvectors of $\mathcal{L}$. That is, similar to
\cite{CSZ06,FWT10}, we significantly reduce the dimension of
$\mathbf{f}$ by requiring it to take the form of $\mathbf{f} =
V_{m}\alpha$ where $V_{m}$ is an $n \times m$ matrix whose columns
are the $m$ eigenvectors with smallest eigenvalues (i.e. the first
$m$ columns of $V$), which can simultaneously ensure that
$\mathbf{f}$ is as smooth as possible in terms of our $L_1$-norm
smoothness. According to equation (\ref{eq:l1ssl}), the objective
function of our $L_1$-SSL can now be formulated as follows:
\begin{eqnarray}
\tilde{Q}(\alpha)&=& \frac{1}{2} ||(V_{m}\alpha)-\mathbf{y}||_2^2 +
\lambda ||(\Sigma^{\frac{1}{2}}V^T)(V_{m}\alpha)||_1 \nonumber \\
&=& \frac{1}{2} ||V_{m}\alpha-\mathbf{y}||_2^2+
\lambda||\sum_{i=1}^m \Sigma^{\frac{1}{2}}(V^T
V_{.i}) \alpha_i||_1 \nonumber \\
&=& \frac{1}{2} ||V_{m}\alpha-\mathbf{y}||_2^2+ \lambda \sum_{i=1}^m
\Sigma^{\frac{1}{2}}_{ii} |\alpha_i|.
\end{eqnarray}
The first term of $\tilde{Q}(\alpha)$ denotes the linear
reconstruction error, while the second term denotes the weighted
$L_1$-norm sparsity regularization over the reconstruction
coefficients. That is, our $L_1$-norm semi-supervised learning has
successfully been transformed into a generalized sparse coding
problem.

\begin{algorithm}[tb]
   \caption{Fast $L_1$-SSL Algorithm}
   \label{alg:l1ssl}
\begin{algorithmic}
   \STATE {\bfseries Input:} the initial label vector $\mathbf{y}$, the weight matrix $W$ of the $k$-NN graph,
                             the number of smallest eigenvectors $m$, and the regularization parameter $\lambda$
   \STATE {\bfseries Output:} the predicted labels by $\mathrm{sign}(\mathbf{f}^*)$
   \STATE Step 1. Compute the normalized Laplacian matrix $\mathcal{L}= I-D^{-\frac{1}{2}}WD^{-\frac{1}{2}}$, where
                  $D=\mathrm{diag}\{\sum_j W_{ij}\}$.
   \STATE Step 2. Find the $m$ smallest eigenvectors of $\mathcal{L}$ stored in $V_m$.
   \STATE Step 3. Solve the $L_1$-norm optimization problem $\alpha^*=\arg\min_{\alpha}
                  \tilde{Q}(\alpha)$ using the modified FISTA.
   \STATE Step 4. Compute $\mathbf{f}^* = V_m \alpha^*$.
\end{algorithmic}
\end{algorithm}

\begin{figure*}[t]
\begin{center}
\subfigure[]{
\includegraphics[height=0.24\textwidth]{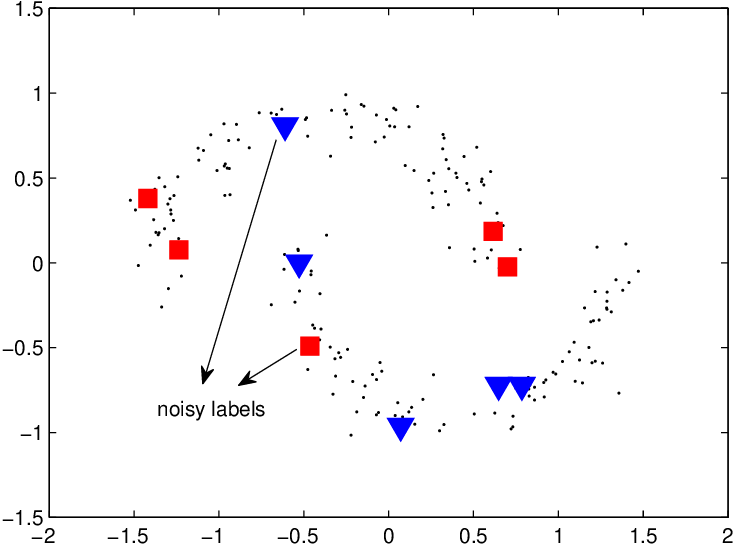}
\label{Fig.2:a}} \hspace{0.4in} \subfigure[]{
\includegraphics[height=0.24\textwidth]{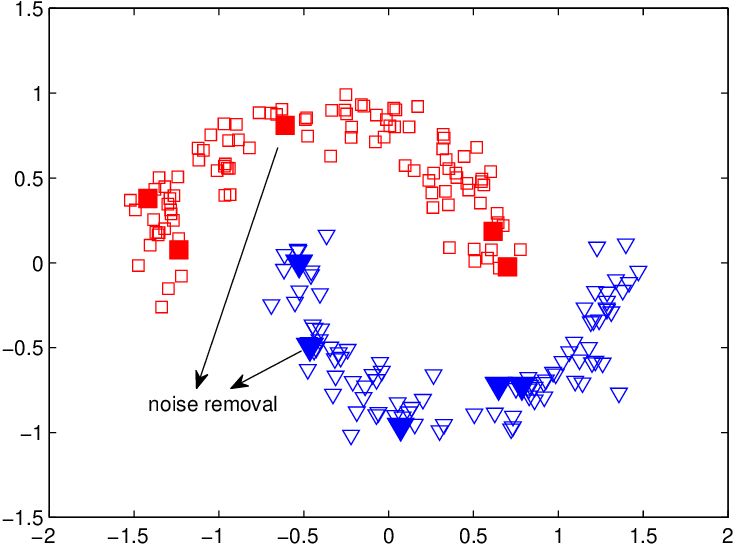}
\label{Fig.2:b}}
\end{center}
\vspace{-0.1in} \caption{(a) The two-moons toy dataset with each
class having one incorrectly (out of five) labeled data point
initially. (b) The classification results on this toy dataset by our
fast $L_1$-norm semi-supervised learning algorithm. Different from
the traditional semi-supervised leaning, the negative effect of
noisy labels can be completely suppressed by our new $L_1$-norm
semi-supervised learning when noisy initial labels are provided.}
\label{Fig.2} \vspace{-0.03in}
\end{figure*}

The formulation $\mathbf{f} = V_{m}\alpha$ used in equation (11) has
two distinct advantages. Firstly, we can derive a linear
reconstruction problem from the original semi-supervised learning
problem, and correspondingly we can \emph{explain our $L_1$-norm
semi-supervised learning in the framework of sparse coding}. This
also provides further insight into Laplacian regularization. In
fact, the second term of $\tilde{Q}(\alpha)$ corresponds to both
Laplacian regularization and sparsity regularization. By unifying
these two types of regularization, we thus successfully obtain novel
$L_1$-norm semi-supervised learning. Secondly, since
$\tilde{Q}(\alpha)$ is minimized with respect to $\alpha \in R^m$
($m\ll n$), we can readily develop fast sparse coding algorithms for
our $L_1$-norm semi-supervised learning. That is, although many
sparse coding algorithms scale polynomially with $m$, they have
linear time complexity with respect to the data size $n$. More
importantly, we \emph{have eliminated the need to compute the full
matrix $B$} in equation (\ref{eq:l1ssl}), which is especially
suitable for image analysis on large datasets. In fact, we only need
to compute the $m$ smallest eigenvectors of $\mathcal{L}$. To speed
up this step, we construct $k$-NN graphs for our $L_1$-norm
semi-supervised learning. Given a $k$-NN graph ($k\ll n$), the time
complexity of finding $m$ smallest eigenvectors of sparse
$\mathcal{L}$ is $O(m^3+m^2n+kmn)$, which is scalable with respect
to the data size $n$.

In theory, any fast sparse coding algorithm can be adopted to solve
the $L_1$-norm optimization problem $\min_{\alpha}
\tilde{Q}(\alpha)$. In this paper, we only consider the Fast
Iterative Shrinkage-Thresholding Algorithm (FISTA) \cite{BT09},
since its implementation mainly involves lightweight operations such
as vector operations and matrix-vector multiplications. To adjust
the original FISTA for our $L_1$-norm semi-supervised learning, we
only need to modify the soft-thresholding function as:
\begin{eqnarray}
\mathrm{soft}(\alpha_i,\frac{\lambda \Sigma^{\frac{1}{2}}_{ii}}
{||V_m||_s^2}) = \mathrm{sign}(\alpha_i)
\max\{|\alpha_i|-\frac{\lambda \Sigma^{\frac{1}{2}}_{ii}}
{||V_m||_s^2}, 0\},
\end{eqnarray}
where $||V_m||_s$ represents the spectral norm of the matrix $V_m$.
For large problems, it is often computationally expensive to
directly compute the Lipschitz constant $||V_m||_s^2$. In practice,
it can be efficiently estimated by a backtracking line-search
strategy \cite{BT09}. The complete algorithm for our fast $L_1$-norm
semi-supervised learning is outlined in Algorithm~\ref{alg:l1ssl}.
Since both Step 2 and Step 3 are scalable with respect to the data
size $n$, our algorithm can be applied to large problems.

\begin{figure}[t]
\vspace{0.02in}
\begin{center}
\includegraphics[width=0.41\textwidth]{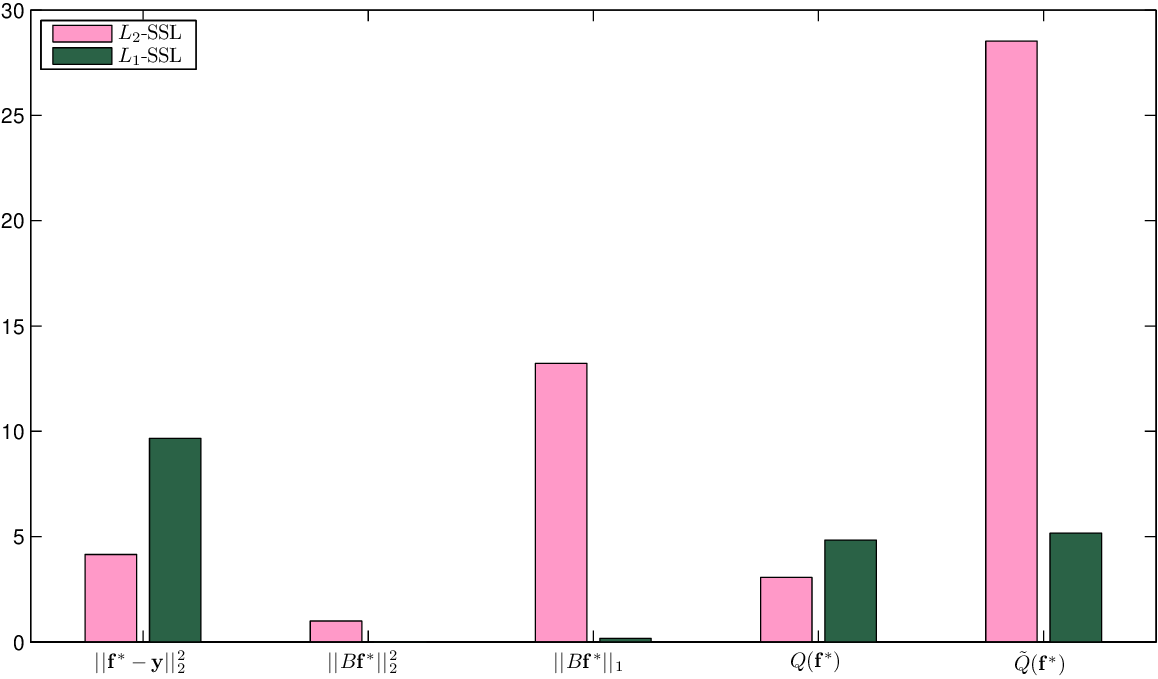}
\end{center}
\vspace{-0.1in} \caption{The quantitative comparison between
$L_2$-SSL and $L_1$-SSL on the two-moons toy dataset given by
Fig.~\ref{Fig.2:a}. Here, $\mathbf{f}^*$ denotes the best solution
found by $L_2$-SSL or $L_1$-SSL when $\lambda$ is set to the same
value. The best solution found by our $L_1$-SSL is shown to be
extremely smooth no matter which smoothness measure is considered,
which is not true for $L_2$-SSL.} \label{Fig.3} \vspace{-0.15in}
\end{figure}

The classification results by our $L_1$-SSL algorithm on the
two-moons toy dataset are shown in Fig.~\ref{Fig.2}. We find that
our algorithm can handle the problem (see Fig.~\ref{Fig.1:b})
associated with the traditional $L_2$-SSL when noisy labels are
provided initially. That is, our $L_1$-SSL algorithm can benefit
from the nice property of sparsity induced by $L_1$-norm
optimization and thus effectively suppress the negative effect of
noisy labels. To give more convincing verification of such
noise-robustness advantage, we further show a quantitative
comparison between $L_2$-SSL and $L_1$-SSL in Fig.~\ref{Fig.3}. The
best solution $\mathbf{f}^*$ found by $L_2$-SSL or $L_1$-SSL is
evaluated here by five quantitative measures such as the fitting
error $||\mathbf{f}^*-\mathbf{y}||_2^2$, the traditional smoothness
$||B\mathbf{f}^*||_2^2$, and the $L_1$-norm smoothness
$||B\mathbf{f}^*||_1$.

We can clearly observe from Fig.~\ref{Fig.3} that the best solution
$\mathbf{f}^*$ found by $L_2$-SSL is not smooth in terms of
$||B\mathbf{f}^*||_1$, although smooth in terms of
$||B\mathbf{f}^*||_2^2$. That is, considering the $L_1$-norm
smoothness measure, we have indirectly shown that $L_2$-SSL can be
severely misled by noisy labels (also consistent with
Fig.~\ref{Fig.1}). Here, it is worth noting that the less smooth a
solution is, the poorer its generalization ability is (and thus more
possible to be misled by the noise). On the contrary, the best
solution $\mathbf{f}^*$ found by our $L_1$-SSL is shown to be
extremely smooth no matter which smoothness measure is considered.
Hence, by simultaneously controlling the fitting error below a low
level, our $L_1$-SSL has successfully suppressed the negative effect
of noisy labels (also consistent with Fig.~\ref{Fig.2}).

\section{Applications to Robust Image Analysis}
\label{sect:ria}

\begin{figure*}[t]
\vspace{0.05in}
\begin{center}
\includegraphics[width=0.90\textwidth]{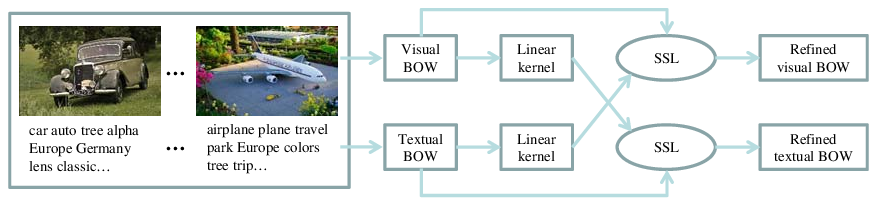}
\end{center}
\vspace{-0.10in} \caption{The flowchart of sparse co-refinement of
visual and textual BOW models by our $L_1$-norm semi-supervised
learning only with linear kernel.} \label{Fig.4} \vspace{-0.1in}
\end{figure*}

Considering the distinct advantage (i.e. noise robustness) of the
proposed $L_1$-SSL algorithm, we apply it to two challenging tasks
of robust image analysis: noise-robust semi-supervised image
classification and noise reduction for both visual and textual BOW
models. Although only tested in these two applications, the proposed
$L_1$-SSL algorithm can be similarly extended to other image
analysis tasks, since semi-supervised learning has been widely used
in the literature.

\subsection{Noise-Robust Semi-Supervised Image Classification}

As the basis of many image analysis tasks such as image annotation
and retrieval, semi-supervised image classification has been
extensively studied in the literature
\cite{XY09,GVS10,LI10,FWT10,TYH09}. In these applications, the
manual labeling of training data is often tedious and expensive,
while the access to unlabeled data is much easier. The original
motivation of semi-supervised image classification is just to reduce
the need for expensive labeled data by exploiting the large number
of unlabeled data.

In this paper, we consider a more challenging problem, i.e.,
semi-supervised image classification with both correctly and
incorrectly labeled data. In general, the occurrence of noisy labels
may be due to the subjective manual labeling of training data.
Fortunately, this challenging problem can be addressed to some
extent by our $L_1$-SSL algorithm. As we have mentioned, our
$L_1$-SSL algorithm can benefit from the nice property of sparsity
induced by $L_1$-norm optimization and thus effectively suppress the
negative effect of noisy labels. Since we focus on providing
convincing verification of this noise-robustness advantage, we
directly apply our $L_1$-SSL algorithm to semi-supervised image
classification with noisy initial labels, without considering any
preprocessing or postprocessing techniques. Hence, we only need to
extend Algorithm 1 to multi-class problems commonly encountered in
image analysis, which is elaborated in the following.

We first formulate a multi-class semi-supervised classification
problem the same as \cite{ZBLW04}. Given a dataset $\mathcal{X} =
\{x_1, ..., x_l, x_{l+1},..., x_n\}$ and a label set $\{1,...,C\}$
($C$ is number of classes), the first $l$ data points $x_i~(i\leq
l)$ are labeled as: $y_{ij} = 1$ if $x_i$ belongs to class $j$
($1\leq j \leq C$) and $y_{ij} = 0$ otherwise, while the remaining
data points $x_u~(l+1 \leq u \leq n)$ are unlabeled with $y_{uj} =
0$. The goal of semi-supervised classification is to predict the
labels of the unlabeled data points, i.e., to find a matrix $F
=[f_{ij}]_{n\times C}$ corresponding to a classification on the
dataset $\mathcal{X}$ by labeling each data point $x_i$ with a label
$\arg \max_{1\leq j \leq C} f_{ij} $. Let $Y=[y_{ij}]_{n\times C}$,
and we can readily observe that $Y$ is exactly consistent with the
initial labels according to the decision rule. When noisy initial
labels are provided for semi-supervised classification, some entries
of $Y$ may be inconsistent with the ground truth.

Based on the above preliminary notations, we further formulate our
multi-class $L_1$-SSL problem as:
\begin{eqnarray}
\min_F\tilde{Q}(F)= \min_F \frac{1}{2}||F-Y||_F^2 + \lambda
||BF||_1,
\end{eqnarray}
which can be decomposed into $C$ independent subproblems:
\begin{eqnarray}
\min_{F_{.j}}\tilde{Q}(F_{.j})= \min_{F_{.j}}
\frac{1}{2}||F_{.j}-Y_{.j}||_2^2 + \lambda||BF_{.j}||_1,
\end{eqnarray}
where $F_{.j}$ and $Y_{.j}$ denote the $j$-th column of $F$ and $Y$,
respectively. Since each subproblem $\min_{F_{.j}}\tilde{Q}(F_{.j})$
can be regarded as a two-class problem, we can readily solve it by
Algorithm 1. Let $F_{.j}^*= \arg\min_{F_{.j}}\tilde{Q}(F_{.j})$, and
we can classify $x_i$ into class $\arg \max_{1\leq j \leq C}
f_{ij}^*$.

\subsection{Sparse Co-Refinement of Visual and Textual BOW Models}
\label{sect:ria:bowr}

We further pay attention to visual and textual BOW refinement to
obtain robust image representation, which is different from
semi-supervised image classification as a high-level semantic
analysis task. Although both visual and textual BOW models have been
shown to achieve impressive results, each BOW model has its own
drawbacks. Firstly, since the visual BOW model generally creates a
visual vocabulary by clustering on the local descriptors extracted
from images, the visual vocabulary may be far from accurate due to
the inherent limitation of clustering and thus the labels of local
descriptors may be rather noisy. This means that visual BOW
refinement is crucial for the success of BOW-based image analysis
tasks. Secondly, instead of the expensive manual labeling of images,
the textual BOW model for image representation is commonly based
upon the image tags contributed by the community (e.g. Flickr) or
automatically derived from the associated text (e.g. Web page).
Because the tags of an image obtained in these ways may be incorrect
and incomplete, the problem of textual BOW refinement becomes rather
challenging.

To address the above problems, we propose a novel framework for
sparse co-refinement of visual and textual BOW models by our
$L_1$-SSL algorithm, as shown in Fig.~\ref{Fig.4}. Our basic idea is
to formulate BOW refinement as a multi-class semi-supervised
learning (SSL) problem by regarding each word of the BOW model as a
``class" so that our noise-robust $L_1$-SSL algorithm can be applied
to noise reduction for both visual and textual BOW models. Since
textual BOW refinement is actually a dual problem of visual BOW
refinement, we focus on visual BOW refinement in the following.

Let $Y \in R^{n\times M}$ be the visual BOW representation and $B
\in R^{n\times n}$ be computed based on the textual BOW
representation, where $n$ is the number of images and $M$ is the
number of visual words. To compute $B$ according to equation (6), we
only consider a linear kernel matrix (used as the weight matrix $W$)
defined with the textual BOW representation. The visual BOW
refinement problem can be formulated as:
\begin{eqnarray}
\min_F \frac{1}{2}||F-Y||_F^2 + \lambda ||BF||_1 + \gamma||F-Y||_1,
\end{eqnarray}
where $\lambda$ and $\gamma$ denote two regularization parameters.
As compared to the $L_1$-SSL problem given by equation (13), the
only difference is that another $L_1$-norm regularization term (i.e.
$||F-Y||_1$) is considered for visual BOW refinement.

As we have mentioned in Section~\ref{sect:l1ssl:l1lap}, we do not
formulate the fitting constraint as an $L_1$-norm term for our
$L_1$-SSL because the predicted labels will be too sparse when very
few labeled data are provided initially. As a truth, the sparsity of
predicted labels completely conflicts with the original goal of
semi-supervised learning. However, the case is quite different for
visual BOW refinement, i.e., a large number of initial labeled data
are provided since each visual word can be assigned to many images.
Hence, the predicted labels may not be sparse even if the $L_1$-norm
fitting constraint is used for semi-supervised learning. Here, our
main motivation of considering $||F-Y||_1$ is to induce the fitting
error sparsity and thus impose direct noise reduction on $Y$.

Although we can find a unique global solution for the visual BOW
refinement problem by convex optimization, it is not easy to develop
an efficient algorithm for this convex optimization. Fortunately, we
can approximately solve it in two $L_1$-norm optimization steps: (1)
$Y^*=\arg\min_F \frac{1}{2}||F-Y||_F^2 + \lambda ||BF||_1$; (2) $F^*
= \arg\min_F \frac{1}{2}||F-Y^*||_F^2 + \gamma||F-Y||_1$. The first
optimization subproblem can be efficiently solved by our $L_1$-SSL
algorithm, while the second subproblem has an explicit solution
based on the soft-thresholding function:
\begin{eqnarray}
F^*=\mathrm{soft}(Y^*-Y,\gamma)+Y,
\end{eqnarray}
where the definition of $\mathrm{soft}(\cdot,\cdot)$ can be found in
equation (12). Considering the scalability of our $L_1$-SSL
algorithm with respect to the data size, the visual BOW refinement
problem can be solved in a linear time cost.

As a dual problem, the textual BOW refinement can be formulated in
the same form of equation (15) by computing $B$ using the visual BOW
representation instead. In summary, we have successfully solve the
challenging problem of sparse co-refinement of visual and textual
BOW models based on our $L_1$-SSL algorithm. To our best knowledge,
we have made the first attempt to obtain robust image representation
by sparse co-refinement of visual and textual BOW models, which is
extremely important for the success of robust image analysis on
community-contributed image collections (e.g. Flickr). However, in
the literature, most previous methods can not deal with visual and
textual BOW refinement simultaneously. For example, various
supervised \cite{YJS08,MJJ10} and unsupervised \cite{JXY09,LPI11}
methods have been developed specially for visual vocabulary
optimization, while in \cite{LNR10,ZYM10,TYH09} only tag refinement
is considered for robust image analysis.

It is worth noting that the supervisory information is usually
expensive to obtain for visual vocabulary optimization in
\cite{YJS08,MJJ10}, while the access to the image tags used for our
visual BOW refinement is much easier (although noisy). Moreover, the
use of image tags also distinguishes our visual BOW refinement
method from the unsupervised methods \cite{JXY09,LPI11} without
considering any high-level semantic information for visual
vocabulary optimization. As compared to the closely related work
\cite{TYH09} on tag refinement that only adopts the traditional
Laplacian regularization for semi-supervised learning, this paper
has formulated new $L_1$-norm Laplacian regularization which has a
wide and important use in the literature.

\section{Experimental Results}
\label{sect:exp}

In this section, our $L_1$-SSL algorithm is tested in two
applications: noise-robust image classification and co-refinement of
visual and textual BOW models. In particular, to show the
descriptive power of the refined BOW models, we apply them to
supervised classification with SVM, different from semi-supervised
classification in the first application.

\subsection{Noise-Robust Image Classification}

We evaluate our $L_1$-SSL algorithm for noise-robust image
classification on the four image datasets listed in
Table~\ref{Table.1}. We first describe the experimental setup and
then compare our $L_1$-SSL algorithm with other closely related
methods.

\subsubsection{Experimental Setup}

\begin{table}[t]
\vspace{0.05in} \caption{Details of the four image datasets
including two handwritten digit datasets and two natural image
datasets.} \label{Table.1} \vspace{-0.10in}
\begin{center}
\tabcolsep0.36cm
\begin{tabular}{|c|cccc|}
\hline
Datasets      & MNIST  &  USPS  &  Corel  &  Scene \\
\hline
\#samples    & 10,000 &  9,298 &  2,000  &  2,688 \\
\#features   & 784    &  256   &  400    &  400   \\
\#classes    & 10     &  10    &  20     &  8     \\
\hline
\end{tabular}
\end{center}
\vspace{-0.10in}
\end{table}

Our $L_1$-norm semi-supervised learning ($L_1$-SSL) is compared to
four other representative methods: (1) the traditional $L_2$-norm
semi-supervised learning ($L_2$-SSL) \cite{ZBLW04}, (2) Lasso-based
$L_1$-norm semi-supervised learning (Lasso-SSL) \cite{SB10}, (3)
linear neighborhood propagation (LNP) \cite{WZ08}, and (4) support
vector machine (SVM). To make an extensive comparison, we conduct
two groups of experiments: semi-supervised classification with a
varying number of clean initial labeled images, and noise-robust
classification with a varying percentage of noisy initial labeled
images. The test accuracies on the unlabeled images are averaged
over 25 independent runs and used for performance evaluation.

\begin{figure*}[t]
\vspace{0.05in}
\begin{center}
\includegraphics[width=0.92\textwidth]{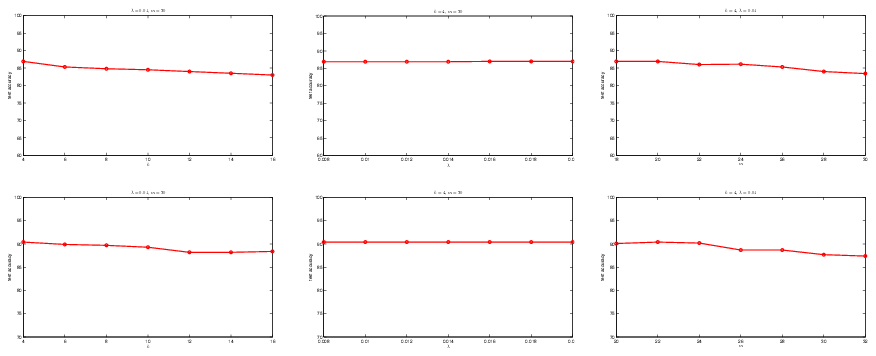}
\end{center}
\vspace{-0.10in} \caption{Illustration of the effect of different
parameters on our $L_1$-SSL algorithm with 50 clean initial labeled
images for the two handwritten digit datasets. \textbf{First Row}:
MNIST. \textbf{Second Row}: USPS.} \label{Fig.5} \vspace{-0.10in}
\end{figure*}

\begin{figure*}[t]
\vspace{0.05in}
\begin{center}
\includegraphics[width=0.79\textwidth]{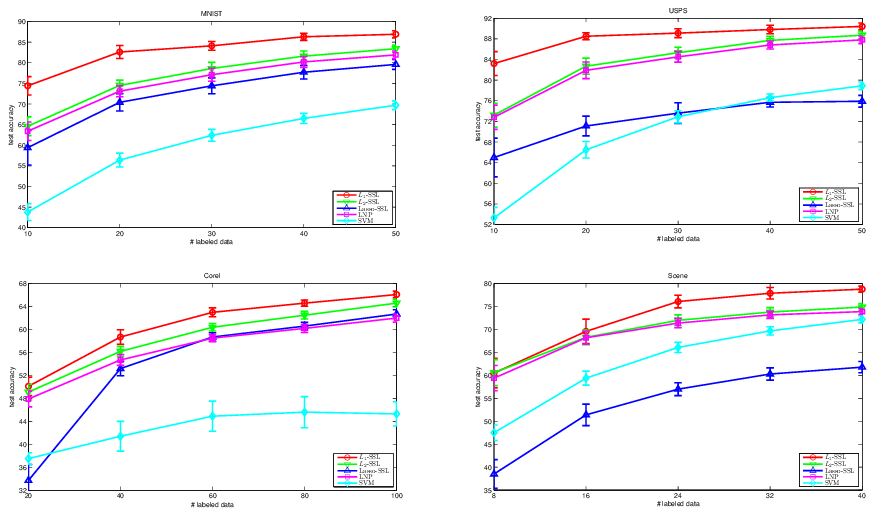}
\end{center}
\vspace{-0.1in} \caption{The classification results (\%) on the four
image datasets by different algorithms when a varying number of
clean labeled images are initially provided. The error bar indicates
the 95\% confidence interval.} \label{Fig.6} \vspace{-0.1in}
\end{figure*}

We adopt two different approaches to kernel matrix computation.
Firstly, for the two handwritten digit datasets (i.e. MNIST and
USPS), we compute the Gaussian kernel matrix according to equation
(1) with fixed $\sigma=1$. Secondly, for the two natural image
datasets (i.e. Scene and Corel), we compute the spatial Markov
kernel matrix \cite{LI11} based on 400 visual words (i.e. 400
features), just the same as \cite{LI10ECCV}. The kernel matrix can
be directly used for SVM, while for semi-supervised learning we can
regard it as the weight matrix so that a $k$-NN graph can be
constructed. The $k$-NN graph is further refined for LNP by
quadratic programming \cite{WZ08}.

We empirically select $k=4$, $\lambda=0.01$ and $m=20$ for our
$L_1$-SSL algorithm on the two handwritten digit datasets. Here, it
should be noted that both $k$ and $m$ are determined by the
consideration of the tradeoff between running efficiency and
classification performance, i.e., we always prefer smaller $k$ and
$m$ for our $L_1$-SSL algorithm when there only exists little
performance degradation. More importantly, as shown in
Fig.~\ref{Fig.5}, our $L_1$-SSL algorithm is generally not much
sensitive to these parameters. The same strategy of parameter
selection is adopted for our $L_1$-SSL algorithm on the two natural
image datasets, while the parameters of other related algorithms for
comparison are also set their respective optimal values.

\subsubsection{Classification Results}

\begin{figure*}[t]
\vspace{0.05in}
\begin{center}
\includegraphics[width=0.79\textwidth]{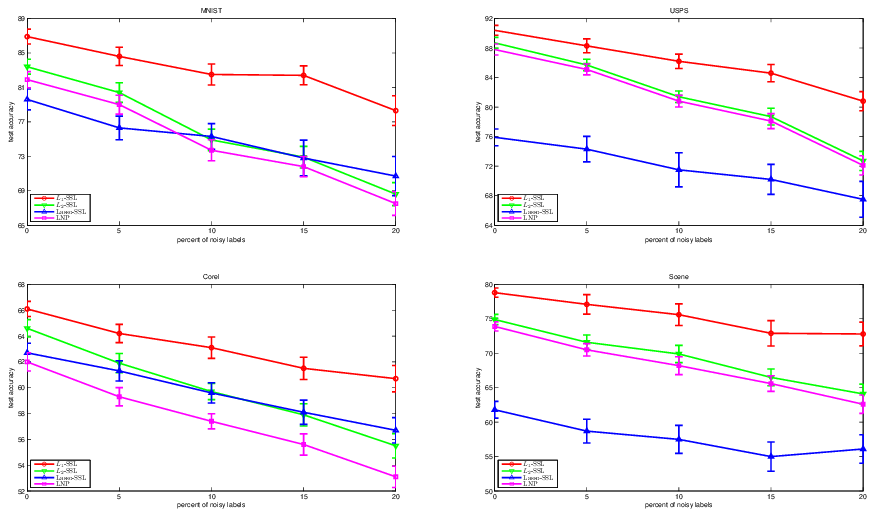}
\end{center}
\vspace{-0.12in} \caption{The classification results (\%) on the
four image datasets by different algorithms when a varying
percentage of noisy labels are provided (among totally
$5\times$\#classes initial labeled images). The error bar indicates
the 95\% confidence interval.} \label{Fig.7} \vspace{-0.15in}
\end{figure*}

Although our original motivation is to apply our $L_1$-SSL to
noise-robust classification, we first compare the five different
methods in the less challenging task of semi-supervised
classification with clean initial labeled images to verify their
effectiveness in dealing with the scarcity of labeled images. The
comparison results are shown in Fig.~\ref{Fig.6}, where the 95\%
confidence intervals are also provided. In general, we can observe
that our $L_1$-SSL consistently performs the best among all the five
methods. The reason may be that our $L_1$-SSL can benefit form the
sparsity induced by our $L_1$-norm Laplacian regularization and thus
suppress the negative effect of the complicated manifold structure
hidden among images on semi-supervise classification. It should be
noted that the four image datasets have much more complicated
structures than the two-moons toy dataset shown in
Fig.~\ref{Fig.1:a}, which are really challenging to deal with for
the other four methods including $L_2$-SSL. Interestingly, although
an $L_1$-norm semi-supervised learning strategy is also adopted,
Lasso-SSL \cite{SB10} ignores the important Laplacian regularization
and thus generally performs the worst among the four SSL methods.

We make further comparison in the challenging task of noise-robust
classification with noisy initial labeled images. Since the four SSL
methods have been shown to generally outperform SVM (see
Fig.~\ref{Fig.6}), we focus on verifying the effectiveness of noise
reduction by semi-supervised learning in the following. The
comparison results on noise-robust classification are shown in
Fig.~\ref{Fig.7}. We find that our $L_1$-SSL consistently achieves
significant gains over the other SSL methods, especially when more
noisy labels are provided initially. That is, our $L_1$-norm
Laplacian regularization indeed can help to find a smooth and also
sparse solution for semi-supervised learning and thus effectively
suppress the negative effect of noisy labels. More importantly,
although all the four methods suffer from more performance
degradation when the percentage of noisy labels grows, the
performance of $L_1$-SSL and Lasso-SSL degrades the slowest due to
that they both utilize sparse coding for semi-supervised learning.

Besides the above advantages, our $L_1$-SSL has another advantage in
terms of running efficiency, i.e., it runs the fastest among the
four SSL algorithms. For example, the time taken by $L_1$-SSL,
$L_2$-SSL, Lasso-SSL, and LNP on the MNIST dataset with 50 clean
labeled images is 39, 57, 433, and 132 seconds, respectively. We run
all the algorithms (Matlab code) on a server with 3GHz CPU and
31.9GB RAM.

\begin{figure*}[t]
\vspace{0.05in}
\begin{center}
\includegraphics[width=0.92\textwidth]{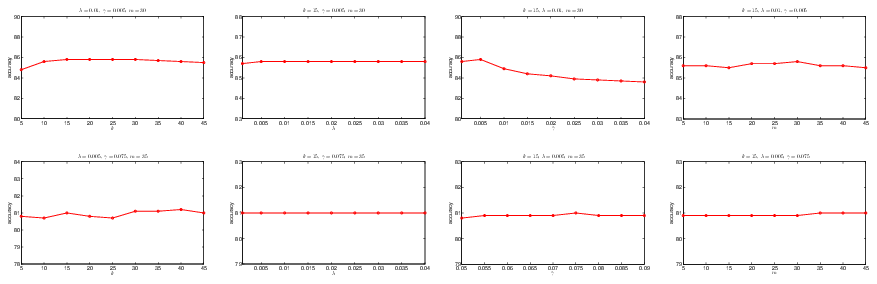}
\end{center}
\vspace{-0.12in} \caption{The twofold cross-validation results by
our $L_1$-SSL algorithm on the training set of the Flickr image
dataset. \textbf{First Row}: visual BOW refined with textual BOW.
\textbf{Second Row}: textual BOW refined with visual BOW.}
\label{Fig.8} \vspace{-0.15in}
\end{figure*}

\subsection{Visual and Textual BOW Refinement}

In this subsection, our $L_1$-SSL algorithm is applied to sparse
co-refinement of visual and textual BOW models. To verify the
descriptive power of the refined BOW models, we focus on evaluating
them in SVM-based image classification. Here, it should be noted
that the refined BOW models can also be readily extended to many
other image analysis tasks such as content-based and text-based
image retrieval. Moreover, although the visual and textual BOW
refinement problems can be solved by any SSL method (see
Fig.~\ref{Fig.4}), we only make comparison between our $L_1$-SSL and
$L_2$-SSL, since they have both been shown to generally outperform
the other SSL methods in the above experiments.

\subsubsection{Experimental Setup}

We conduct a group of experiments on a Flickr benchmark dataset
\cite{FIL11}, which consists of totally 8,564 images crawled from
the photo sharing website Flickr. This image dataset is organized
into eleven categories: airplane, auto, dog, turtle, elephant, NBA,
laptop, piano, farm, cityscape and library. The high-level category
labels of images can be used for SVM-based image classification. In
the following experiments, we split this dataset into a training set
of 4,282 images and a test set of the same size.

To obtain the visual BOW representation for the Flickr dataset, we
extract the SIFT descriptors of $16\times 16$ pixel blocks computed
over a regular grid with spacing of 8 pixels. We then perform
$k$-means clustering on the extracted descriptors to form a
vocabulary of 2,000 visual words. Here, we aim to make the visual
BOW representation more noisy by considering a relatively larger
visual vocabulary. Moreover, we generate the textual BOW
representation based on the user-provided textual tags. As a
preprocessing step, we remove the stop words and check the remaining
tags against the WordNet to remove the tags that do not exist. The
final textual vocabulary only contains the most frequent 1,000
words.

From these two BOW representations, we only derive linear kernels
for our $L_1$-SSL algorithm in the tasks of visual and textual BOW
refinement. Based on the obtained kernel matrices, we can further
construct $k$-NN graphs for semi-supervised learning. The linear
kernels are also used for the subsequent SVM classification.
According to the twofold cross-validation results on the training
set as shown in Fig.~\ref{Fig.8}, we set the parameters of our
$L_1$-SSL algorithm to their respective optimal values listed in
Table~\ref{Table.2}. Just as what we have done in noise-robust image
classification, we still determine both $k$ and $m$ by the
consideration of the tradeoff between running efficiency and
classification performance. More importantly, we can clearly observe
from Fig.~\ref{Fig.8} that our $L_1$-SSL algorithm is not much
sensitive to these parameters in most cases.

\subsubsection{Refinement Results}

\begin{table}[t]
\vspace{0.05in} \caption{The parameters selected by cross-validation
for our $L_1$-SSL algorithm in both visual and textual BOW
refinement.} \label{Table.2} \vspace{-0.10in}
\begin{center}
\tabcolsep0.45cm
\begin{tabular}{|c|cccc|}
\hline
Parameters     & $k$  & $\lambda$ &  $\gamma$  & $m$\\
\hline
Visual BOW   &  15    &  0.010    &  0.005  & 30 \\
Textual BOW  &  15    &  0.005   &  0.075  & 35 \\
\hline
\end{tabular}
\end{center}
\vspace{-0.15in}
\end{table}

\begin{table}[t]
\vspace{0.05in} \caption{The classification results (\%) on the
Flickr image dataset using different BOW models. Both $L_2$-SSL and
$L_1$-SSL can be used for co-refinement of visual and textual BOW
models.} \label{Table.3} \vspace{-0.10in}
\begin{center}
\tabcolsep0.31cm
\begin{tabular}{|c|cccc|}
\hline
Methods      & Original & \cite{TYH09} & $L_2$-SSL &  $L_1$-SSL  \\
\hline
Visual BOW   & 60.1     & 78.2   &  84.3    &    \textbf{87.4}  \\
Textual BOW  & 77.8     & 81.4   &  81.5    &    \textbf{83.2}  \\
\hline
\end{tabular}
\end{center}
\vspace{-0.15in}
\end{table}

To show the effectiveness of visual and textual BOW refinement, we
compare the refined BOW models by our $L_1$-SSL to: (1) the original
BOW models, (2)  the refined BOW models by the SSL method proposed
in \cite{TYH09}, and (3) the refined BOW models by $L_2$-SSL. The
comparison results are list in Table~\ref{Table.3}. The immediate
observation is that the refined BOW models by our $L_1$-SSL lead to
obvious gains over the original BOW models, especially when the
visual BOW model is refined with the textual BOW model (i.e. 27.3\%
gain). This means that our $L_1$-SSL for visual and textual BOW
refinement indeed can benefit from the sparsity induced by
$L_1$-norm optimization and thus effectively suppress the noise in
both visual and textual BOW models.

Moreover, we can clearly observe from Table~\ref{Table.3} that
$L_2$-SSL also achieves promising results in visual and textual BOW
refinement, although it is not originally developed for noise
reduction. The reason may be that the visual (or textual) words
associated with each image in the Flickr dataset are not only noisy
but also incomplete due to inaccurate clustering (or subjective and
limited manual labeling), while the issue of incomplete words can be
effectively handled by word propagation based on $L_2$-SSL. Here, it
is worth noting that, different from the traditional $L_2$-SSL, our
$L_1$-SSL is suitable for both word propagation and noise reduction.
Hence, as shown in Table~\ref{Table.3}, our $L_1$-SSL performs
better than $L_2$-SSL in both visual and textual BOW refinement.

As for the SSL method \cite{TYH09}, we find that it works nearly as
well as $L_2$-SSL in textual BOW refinement, but leads to much worse
results in visual BOW refinement. Its promising performance in
textual BOW refinement may be due to that it can perform both noise
reduction and word propagation by imposing the fitting error
sparsity on SSL. However, the case is different for visual BOW
refinement, i.e., the issue of incomplete words may be more severe
for wrong label permutation along with inaccurate clustering. As
compared to $L_2$-SSL \cite{ZBLW04} (one of the most outstanding SSL
methods), the SSL method \cite{TYH09} has a poorer performance of
visual word propagation and thus suffers from obvious degradation.

\section{Conclusions}
\label{sect:con}

We have proposed a novel $L_1$-norm semi-supervised learning method
in this paper. Different from the traditional graph-based SSL that
defines Laplacian regularization by a quadratic function, we have
successfully reformulated Laplacian regularization as an $L_1$-norm
term. More importantly, we find that this new formulation is
explicitly based upon the manifold structure of the data. Due to the
resulting $L_1$-norm optimization, our new $L_1$-SSL can benefit
from the nice property of sparsity and thus effectively suppress the
negative effect of noisy labels. Extensive results have shown the
promising performance of our $L_1$-SSL in two challenging tasks of
robust image analysis. In the future work, considering the wide use
of Laplacian regularization, we will apply our new $L_1$-norm
Laplacian regularization to many other challenging problems.
Moreover, the refined visual and textual BOW models by our $L_1$-SSL
will be evaluated in other image analysis tasks such as
content-based and text-based image retrieval.

\section*{Acknowledgements}

The work described in this paper was fully supported by the National
Natural Science Foundation of China under Grant Nos. 60873154 and
61073084.


\begin{IEEEbiography}[{\includegraphics[width=1in,height=1.25in,clip,keepaspectratio]{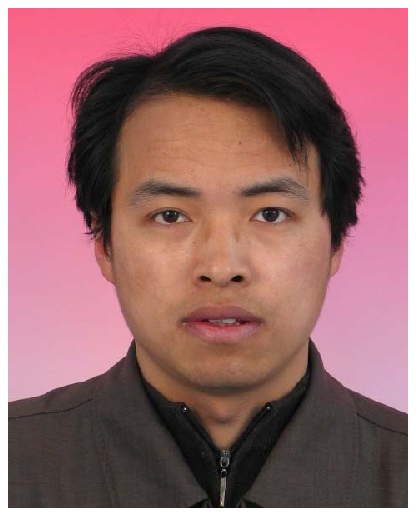}}]{Zhiwu Lu}
received the M.Sc. degree in applied mathematics from Peking
University, Beijing, China in 2005, and the Ph.D. degree in computer
science from City University of Hong Kong in 2011.

Since March 2011, he has become an assistant professor with the
Institute of Computer Science and Technology, Peking University. He
has published over 30 papers in refereed international journals and
conference proceedings including TIP, TSMC-B, TMM, AAAI, ICCV, CVPR,
ECCV, and ACM-MM. His research interests lie in machine learning,
pattern recognition, computer vision, and multimedia information
retrieval.
\end{IEEEbiography}

\begin{IEEEbiography}[{\includegraphics[width=1in,height=1.25in,clip,keepaspectratio]{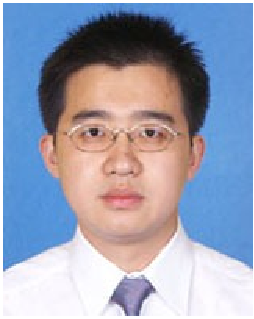}}]{Yuxin Peng}
is the professor and director of Multimedia Information Processing
Lab (MIPL) in the Institute of Computer Science and Technology
(ICST), Peking University. He received his Ph.D. degree in computer
application from School of Electronics Engineering and Computer
Science (EECS), Peking University, in Jul. 2003. After that, he
worked as an assistant professor in ICST, Peking University. From
Aug. 2003 to Nov. 2004, he was a visiting scholar with the
Department of Computer Science, City University of Hong Kong. He was
promoted to associate professor in Peking University in Aug. 2005.
In Aug. 2010, he was promoted to professor in Peking University. He
has published over 50 papers in refereed international journals and
conference proceedings including TCSVT, TIP, ACM-MM, ICCV, CVPR and
AAAI. In 2009, he led his team to participate in TRECVID. In six
tasks of the high-level feature extraction (HLFE) and search, his
team won the first places in fours tasks and the second places in
the left two tasks. Besides, he has obtained 12 patents. His current
research interests include multimedia information retrieval, image
processing, computer vision, and pattern recognition.

\end{IEEEbiography}

\vfill

\end{document}